\documentclass{INTERSPEECH2023}

\interspeechcameraready

\usepackage{cite}
\usepackage{multirow}
\usepackage{threeparttable}
\usepackage{subcaption}

\usepackage{xcolor,colortbl}
\usepackage{hyperref}
\newcommand{\proposed}{Spin}
\newcommand{\transp}{\mathsf{\scriptscriptstyle{T}}}
\usepackage{soul}

\definecolor{LightGreen}{rgb}{0.84,0.96,0.73}

\title{Self-supervised Fine-tuning for Improved Content Representations\\by Speaker-invariant Clustering}
\name{Heng-Jui Chang, Alexander H. Liu, James Glass}
\address{
  MIT CSAIL, USA
}
\email{\{hengjui,alexhliu,glass\}@mit.edu}

\begin{document}

\maketitle
 
\begin{abstract}
Self-supervised speech representation models have succeeded in various tasks, but improving them for content-related problems using unlabeled data is challenging.
We propose speaker-invariant clustering (\proposed), a novel self-supervised learning method that clusters speech representations and performs swapped prediction between the original and speaker-perturbed utterances.
\proposed~disentangles speaker information and preserves content representations with just 45 minutes of fine-tuning on a single GPU.
\proposed~improves pre-trained networks and outperforms prior methods in speech recognition and acoustic unit discovery.\footnote{Code: \url{https://github.com/vectominist/spin}}
\end{abstract}

\section{Introduction}
\label{sec:intro}

Self-supervised learning (SSL) for speech representation using large neural networks and unlabeled data offers effective initialization and representations for downstream tasks~\cite{yang2021superb,evain2021lebenchmark,chang2021exploration,tsai-etal-2022-superb,mohamed2022self}.
Among prior methods, learning discrete units like K-means clusters benefits downstream performance~\cite{hsu2021hubert,chen2022wavlm,chung2021w2v,maekaku2022exploration,chiu2022bestrq,ren2022speech,wells2022phonetic}.
While speech representation encompasses information from multiple aspects, most SSL methods lack explicit speaker disentanglement.
Extracting speaker-invariant linguistic content can benefit downstream tasks like automatic speech recognition~(ASR) and phoneme recognition~(PR)~\cite{hsu2017unsupervised,tjandra2020unsupervised,chan2022content,peyser2022towards,williams2022learning}.
In light of this, ContentVec~\cite{qian2022contentvec} imposes speaker-invariant constraints to pre-trained HuBERT models~\cite{hsu2021hubert} to improve content-related downstream tasks.
However, ContentVec adds a substantial amount of computational cost, requiring 19 hours on 36 GPUs, on top of the pre-trained models, which are already expensive to compute.

In this paper, we first demonstrate the benefits of extracting features closer to the underlying phonetic content to motivate our work.
Following this observation, we present speaker-invariant clustering~(\proposed), a novel and cost-effective self-supervised fine-tuning~(SSFT)\footnote{We use the term SSFT to distinguish fine-tuning methods using only audio~\cite{qian2022contentvec,huang2022improving} from supervised fine-tuning using labeled data~\cite{baevski2020wav2vec2}.} method for SSL models that leverages vector quantization~\cite{caron2018deep,asano2019self,caron2020unsupervised} and speaker disentanglement~\cite{qian2022contentvec} to improve content representation.
In short, \proposed~is trained to identify the unchanged spoken content from pairs of speaker-augmented utterances via quantized representation matching.
Such design leads to a disentangled representation focusing on the spoken content, improving various downstream tasks, including content-related tasks in SUPERB~\cite{yang2021superb} and ZeroSpeech~\cite{nguyen2020zero}.
In terms of efficiency, we show that \proposed~requires less than 45 minutes of training on a single GPU, costing less than 1\% of ContentVec.

\section{Method}
\label{sec:method}

\begin{figure}[t]
    \centering
    \includegraphics[width=0.95\linewidth]{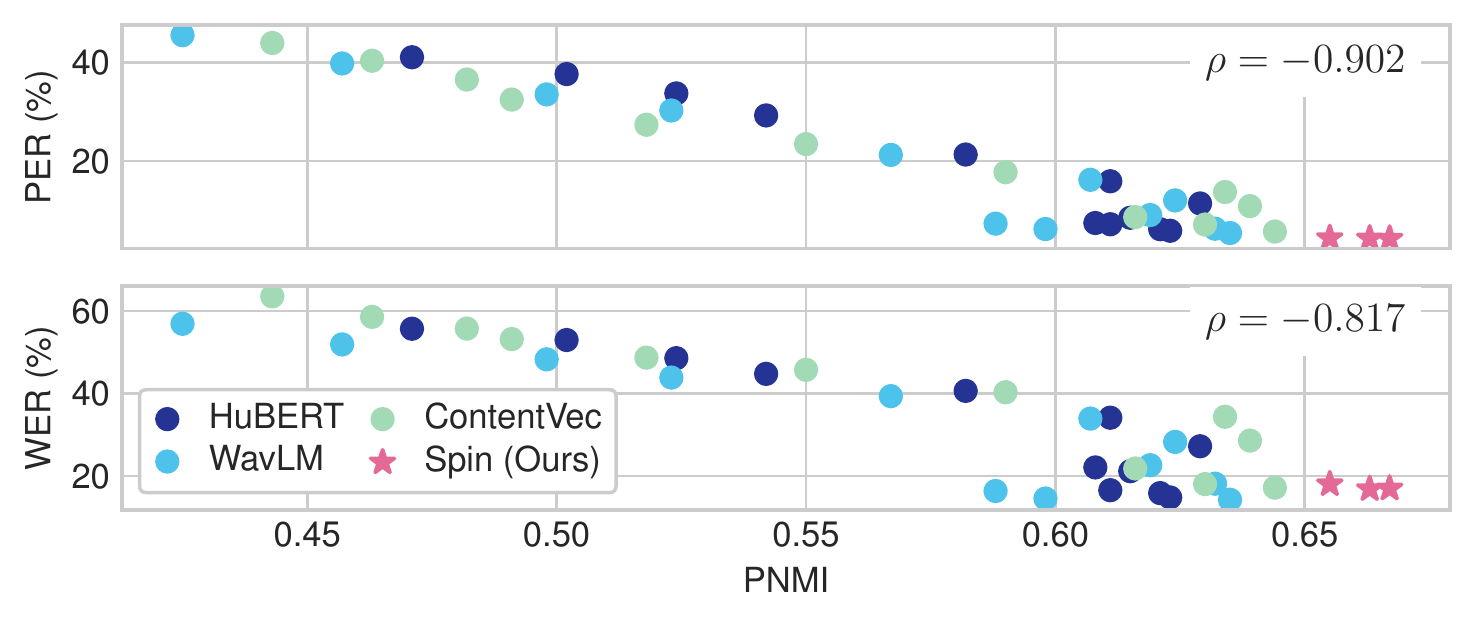}
    \vspace{-14pt}
    \caption{
        Content representation quality (PNMI) vs. phoneme/ word error rates (PER/WER) of SSL model hidden layer representations under a simplified setup in SUPERB~\cite{yang2021superb}.
        $\rho$ is Spearman's rank correlation coefficient.
    }
    \label{fig:corr}
    \vspace{-12pt}
\end{figure}
\begin{figure*}[t]
    \centering
    \includegraphics[width=0.82\linewidth]{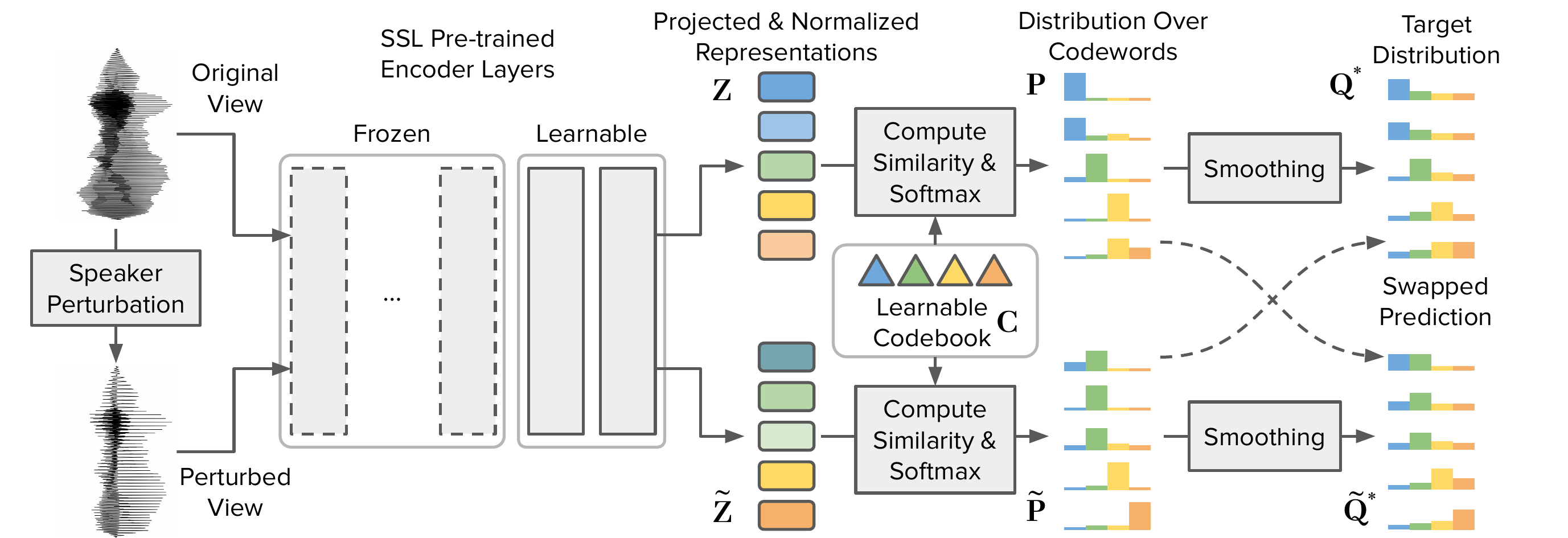}
    \vspace{-10pt}
    \caption{
        The \proposed~architecture.
        A new view is generated with a simple speaker perturbation.
        A pre-trained speech SSL model extracts representations from both utterances ($\mathbf{Z}$/ $\Tilde{\mathbf{Z}}$).
        Representations are projected, normalized, and quantized with a learnable codebook into probability distributions ($\mathbf{P}$/ $\Tilde{\mathbf{P}}$).
        The distributions are smoothed to enforce full codebook usage ($\mathbf{Q}^{\ast}$/ $\Tilde{\mathbf{Q}}^{\ast}$).
        Finally, each frame's distribution is used to predict the target distribution produced by the other view ($\mathbf{P} \rightarrow \Tilde{\mathbf{Q}}^{\ast}$ and $\Tilde{\mathbf{P}} \rightarrow \mathbf{Q}^{\ast}$).
    }
    \label{fig:framework}
    \vspace{-13pt}
\end{figure*}

\subsection{Importance of Content Representation}
\label{subsec:discrete-benefit}

This work assumes representations closer to the underlying phonetic content yield better performance for speaker-invariant downstream tasks like ASR.
To verify this assumption, we extract speech representations from each layer of three pre-trained SSL models (HuBERT~\cite{hsu2021hubert}, WavLM~\cite{chen2022wavlm}, and ContentVec~\cite{qian2022contentvec}) and compute two metrics: 1) the phone-normalized mutual information (PNMI; Sec.~\ref{subsec:cls-qual}) that measures the similarity between phonemes and the discrete units derived by running K-means clustering on the extracted representation~($K=$ 256);
2) the phone/word recognition error rate using the extracted features and a lightweight predictor as detailed in Sec.~\ref{subsec:superb}.

In Fig.~\ref{fig:corr}, higher PNMI representations generally offer better recognition results across all models and layers.
The Spearman's rank correlation coefficients for PNMI-PER~($-$0.902) and PNMI-WER~($-$0.817) verify the strong correlation between the content encoded and downstream performance, leading us to propose an SSFT method that learns from discrete acoustic units to focus on content encoding.

\subsection{Proposed Method}
\label{subsec:method-overview}

\noindent\textbf{Overview.}
An overview of the proposed \proposed~is illustrated in Fig.~\ref{fig:framework}.
Inspired by Swapping Assignments between Views (SwAV)~\cite{caron2020unsupervised} for image representation learning, our idea is to learn speaker-invariant clusters that capture the same content shared between perturbed speech and the original speech.

\vspace{2pt}
\noindent\textbf{Speaker Perturbation.}
To alter the speaker identity without changing the spoken content, we adopt an algorithm proposed by Choi et al.~\cite{choi2021nansy} as ContentVec~\cite{qian2022contentvec}.
The algorithm randomly and uniformly scales formant frequencies and F0, and random equalization is applied.
Because voice information resides in the formant frequencies and F0~\cite{eide1996parametric}, and the content is stored in the relative ratio between formant frequencies~\cite{stevens1987relational}, this algorithm efficiently alters speakers with little content loss.

\vspace{2pt}
\noindent\textbf{Speaker-invariant Clustering.}
With the speaker-augmented and the original speech pair, we aim to discover the consistent underlying content via speaker-invariant clustering.
As in Fig.~\ref{fig:framework}, the output of the original view from the encoder is linearly projected and L2-normalized to representations $\mathbf{Z} = [\boldsymbol{z}_1 \dots \boldsymbol{z}_B]^{\transp} \in\mathbb{R}^{B\times D}$, where $D$ is the dimension of the representations, and $B$ is the number of frames in a batch.
A probability distribution is computed per frame by taking softmax over the scaled cosine similarity between $\mathbf{Z}$ and a learnable codebook of $K$ codewords $\mathbf{C} = [\boldsymbol{c}_1 \dots \boldsymbol{c}_K]^{\transp} \in \mathbb{R}^{K\times D}$ as
\vspace{-5pt}
\begin{equation}
    p\left( k | \boldsymbol{z}_b \right) = \frac{\exp\left( \boldsymbol{z}_b^\transp \boldsymbol{c}_k / \tau \right)}{\sum_{k'} \exp\left(\boldsymbol{z}_b^\transp \boldsymbol{c}_{k'} / \tau \right)},
    \nonumber
    \vspace{-4pt}
\end{equation}
for $k \in [K]$, $b \in [B]$,\footnote{$[N]$ is defined as $\{1, 2, \dots, N \}$.} $\Vert \boldsymbol{c}_k\Vert_2 = 1$, and $\tau>0$ is a scaling temperature.
We define $q\left( k | \Tilde{\boldsymbol{z}}_b \right)$ the distribution over the same codebook using augmented speech.
To learn speaker-invariant clusters that capture the unchanged content, distributions over the codebook should ideally be similar regardless of the speaker, i.e., minimizing the cross-entropy $-q\left( k | \Tilde{\boldsymbol{z}}_b \right) \log p\left( k | \boldsymbol{z}_b \right)$.

\vspace{2pt}
\noindent\textbf{Smoothing for Full Codebook Usage.}
In practice, minimizing the aforementioned cross-entropy term leads to a trivial solution where all representations are clustered into a single codeword if $q$ is obtained similarly with $p$.
To address the issue, we smooth the target distribution $q$ to encourage higher utilization of the codewords.
Following Asano et al.~\cite{asano2019self}, $q$ is obtained by
\vspace{-3pt}
\begin{equation}
    \mathbf{Q}^{\ast} \in \arg\underset{\mathbf{Q}}{\max} ~\mathrm{Tr} \left( \mathbf{Q} \mathbf{C} \mathbf{Z}^\transp \right) + \varepsilon H \left( \mathbf{Q} \right),
    \label{eq:q_star}
    \vspace{-4pt}
\end{equation}
where $\mathbf{Q}^{\ast} \in [0, 1]^{B\times K}$, $q\left( k | \boldsymbol{z}_b \right) = \mathbf{Q}^{\ast}_{b, k}$, and $H \left( \mathbf{Q} \right) = -\sum_{ij} \mathbf{Q}_{ij} \log \mathbf{Q}_{ij}$ is the entropy.
The optimized variable $\mathbf{Q}$ is constrained so that each row is a probability distribution over the $K$ codewords.
When $\varepsilon = $ 0, $q$ is a categorical distribution and easily collapses to using only one codeword.
When $\varepsilon > $ 0, the entropy term smooths the distribution so that all codewords can be utilized more evenly, whereas a higher $\varepsilon$ leads to a more uniform distribution.
Eq.~\ref{eq:q_star} can be efficiently solved by the Sinkhorn-Knopp algorithm on GPUs~\cite{cuturi2013sinkhorn}.
Note that no gradient is applied to $q$.

\vspace{2pt}
\noindent\textbf{Speaker-invariant Swapped Prediction.}
With the smoothed target distribution $q$, the goal is to perform speaker-invariant swapped prediction by minimizing the cross-entropy loss
\vspace{-5pt}
\begin{equation*}
    -\frac{1}{2B}\sum_b \sum_k \left[ 
    q\left( k | \Tilde{\boldsymbol{z}}_b \right) \log p\left( k | \boldsymbol{z}_b \right) + 
    q\left( k | \boldsymbol{z}_b \right) \log p\left( k | \Tilde{\boldsymbol{z}}_b \right) \right],
    \vspace{-6pt}
\end{equation*}
where the second term emerges from the interchangeability of the role of the augmented and original speech.

This objective encourages the model to produce similar representations at the same position between two different views by learning a codebook encoding speaker-invariant acoustic units.
Since learning fewer parameters reduces computation, and top layers encode phonetic content~\cite{pasad2021layer,chang2022distilhubert,tseng2021mandarin}, we propose fine-tuning some top layers to balance the tradeoff between downstream performance and training computation.
Unlike previous methods, \proposed~does not require random masking, so all frames are utilized and contribute to updating the network.
\proposed~is limited to pre-trained models because only the positional information is learned if the model is trained from scratch.

\section{Experiments}
\label{sec:exp}

\begin{table*}[t]
    \centering
    \caption{
        SUPERB~\cite{yang2021superb} phoneme recognition~(PR), automatic speech recognition~(ASR), keyword spotting~(KS), query-by-example~(QbE), intent classification~(IC), and slot filling~(SF).
        Metrics include accuracy~(Acc\%), phoneme error rate~(PER\%), word error rate~(WER\%), maximum term weighted value~(MTWV), F1 score, and concept error rate~(CER\%).
        PT and SSFT denote pre-training and self-supervised fine-tuning.
        Top-3 best results are \underline{underlined}.
        The number of hours of processed speech is computed with Eq.~\ref{eq:proc-hours}.
    }
    \vspace{-8pt}
    \label{tab:superb}
    \begin{threeparttable}
    \begin{tabular}{lccccccc@{}ccc}
        \toprule
        & \multicolumn{2}{c}{\multirow{2}{*}{\shortstack[c]{~\\Training Processed Speech\\in Hours}}} & \multicolumn{4}{c}{Content} &  & \multicolumn{3}{c}{Semantic} \\
        \cmidrule{4-7}
        \cmidrule{9-11}
        \vspace{-2pt}
        & & & PR & ASR & KS & QbE & & IC & \multicolumn{2}{c}{SF} \\
        \cmidrule{2-3}
        \cmidrule{10-11}
        Method & PT & SSFT & PER$\downarrow$ & WER$\downarrow$ & Acc$\uparrow$ & MTWV$\uparrow$ &  & Acc$\uparrow$ & F1$\uparrow$ & CER$\downarrow$ \\
        \midrule
        wav2vec 2.0~\cite{baevski2020wav2vec2}$^{\spadesuit}$ & 640k & 0 & 5.74 & 6.43 & 96.23 & 0.0233 &  & 92.35 & 88.30 & 24.77 \\
        HuBERT~\cite{hsu2021hubert}$^{\spadesuit}$ & 506k & 0 & 5.41 & 6.42 & 96.30 & 0.0736 &  & 98.34 & 88.53 & 25.20 \\
        WavLM~\cite{chen2022wavlm}$^{\spadesuit}$ & 1439k & 0 & 4.84 & 6.31 & \underline{96.79} & \underline{0.0870} &  & \underline{98.63} & \underline{89.38} & \underline{22.86} \\
        data2vec~\cite{baevski2022data2vec}$^{\spadesuit}$ & 420k & 0 & 4.69 & \underline{4.94} & \underline{96.56} & 0.0576 &  & 97.63 & 88.59 & 25.27 \\
        \midrule
        ContentVec\textsubscript{500}~\cite{qian2022contentvec}$^{\clubsuit}$ & 506k & 76k & \underline{4.54}$^{\diamondsuit}$ & \underline{5.70} & 96.40 & 0.0590 &  & \underline{99.10} & \underline{89.60} & \underline{23.60} \\
        HuBERT + \proposed\textsubscript{256} & 506k & 356 & \underline{4.39} & 6.34 & \underline{96.53} & \underline{0.0912} &  & 98.34 & \underline{89.00} & 24.32 \\
        WavLM + \proposed\textsubscript{256} & 1439k & 356 & \underline{4.18} & \underline{5.88} & 96.20 & \underline{0.0879} &  & \underline{98.52} & 88.84 & \underline{24.06} \\
        \bottomrule
    \end{tabular}
    \begin{tablenotes}[flushleft]
        \item
            $^{\spadesuit}$
            {\footnotesize Source: \url{https://superbbenchmark.org/leaderboard} (as of 3/7/2023).}
            $^{\clubsuit}$
            {\footnotesize Reported in: \cite{qian2022contentvec}}
        \item
            $^{\diamondsuit}$
            {\footnotesize Re-implement for a fair comparison (original: 4.90).}
    \end{tablenotes}
    \end{threeparttable}
    \vspace{-10pt}
\end{table*}

\subsection{Setup}
\label{subsec:setup}

\noindent\textbf{Data.}
\proposed~is trained with the LibriSpeech train-clean 100 hours subset~\cite{Panayotov2015libri}, and we found more data does not improve.

\noindent\textbf{Implementation.}
We applied \proposed~to HuBERT~\cite{hsu2021hubert} and WavLM~\cite{chen2022wavlm}, and only the last two layers are fine-tuned (7M parameters per layer).\footnote{Checkpoints: \url{https://github.com/s3prl/s3prl}}
We set $D=$ 256, $\tau = $ 0.1, $\varepsilon = $ 0.02, and sweep the codebook sizes $K\in$ \{128, 256, 512, 1024, 2048\}.
Each view's mini-batch has at most 256 seconds of speech, corresponding to $B=$ 12.8k frames.
The learning rate is first linearly increased from 0 to 10\textsuperscript{$-$4} for 2.5k updates, then linearly decreased to 10\textsuperscript{$-$6} for 2.5k updates.
The Sinkhorn-Knopp algorithm iterates three times to compute $\mathbf{Q}^{\ast}$ per view.
\proposed~is trained on a single RTX A5000 GPU, each taking 45 minutes.
We select models that are trained with all 5k updates.

\subsection{Speech SSL Models}
\textbf{HuBERT} and \textbf{WavLM} are pre-trained to predict cluster IDs of masked audio frames from clustering MFCC features or hidden representations of pre-trained models.
These models serve as baselines for \proposed.
\textbf{data2vec}~\cite{baevski2022data2vec} is trained to masked-predicting hidden representations of the exponential moving average of the model itself.
We avoid applying \proposed~to data2vec because the phonetic content resides at the bottom layers~(Table~\ref{tab:abx}), requiring fine-tuning many more top layers, and thus increasing computation costs.
\textbf{ContentVec} is a stronger baseline as it is also trained to improve extracting content with speaker disentanglement.
ContentVec learns to mask-predict a pre-trained HuBERT hidden representation K-means clusters.
Based on the number of clusters in the target, there are two versions: ContentVec\textsubscript{100} and ContentVec\textsubscript{500}.
These SSL models share a similar architecture: a 7-layer CNN feature extractor followed by a 12-layer transformer encoder~\cite{vaswani2017attention}, approximately having 95M parameters each.
All models are frozen in evaluation tasks, and continuous transformer encoder hidden representations are used unless otherwise specified.

\subsection{SUPERB}
\label{subsec:superb}

\begin{figure}
    \centering
    \includegraphics[trim=6 0 10 0,clip,width=\linewidth]{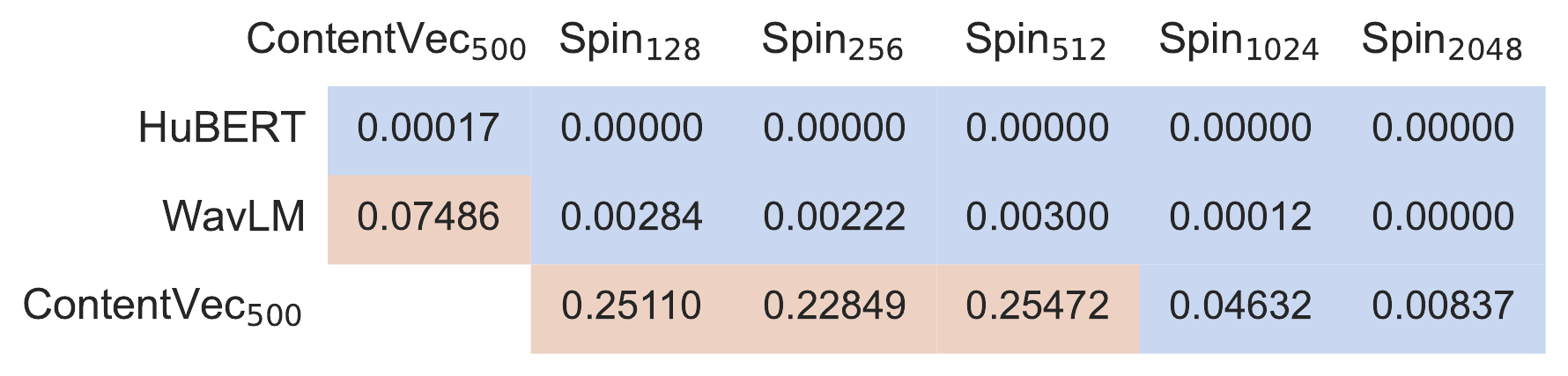}
    \vspace{-22pt}
    \caption{
        t-test $p$-values of SUPERB~\cite{yang2021superb} phoneme recognition error rates.
        All \proposed~models here are based on HuBERT.
    }
    \label{fig:pval}
    \vspace{-10pt}
\end{figure}

This section evaluates \proposed~on content and semantic tasks in the Speech processing Universal PERformance Benchmark (SUPERB)~\cite{yang2021superb}.
Each task and SSL model uses a set of learnable weights to weighted-sum representations across hidden layers of the frozen SSL model.
The aggregated features are then fed to a prediction head for supervised training.
We report phoneme recognition~(PR), automatic speech recognition~(ASR), keyword spotting~(KS), query-by-example spoken term discovery~(QbE), intent classification~(IC), and slot filling~(SF).
We choose $K=$ 256 for \proposed~as it offers the best overall results.

In Table~\ref{tab:superb}, \proposed~benefits learning content representations because HuBERT and WavLM are improved in content-related tasks (PR, ASR, and QbE) while reducing the performance gap with ContentVec.
According to the significance test on PR in Fig.~\ref{fig:pval}, \proposed~passes t-test compared with HuBERT and WavLM.
Increasing the codebook size ($K=$ 1024 and 2048) outperforms ContentVec with a $p < $ 0.05.
Next, we show the hours of processed speech during training
\vspace{-5pt}
\begin{equation}
    \text{processed speech} = \text{training steps} \times \text{effective batch size}
    \label{eq:proc-hours}
\end{equation}
to quantify machine-independent training costs.
Based on these data, \proposed~requires less than 0.5\% computation of ContentVec to outperform in PR and QbE while offering similar performance in other tasks.
Moreover, most models perform similarly in KS and IC, and we found these tasks sensitive to hyperparameters, making them less suitable for comparison.
Overall, \proposed~improves SSL models with a meager budget.

\begin{figure*}
    \centering
    \begin{subfigure}[b]{0.33\linewidth}
         \centering
         \includegraphics[height=100pt]{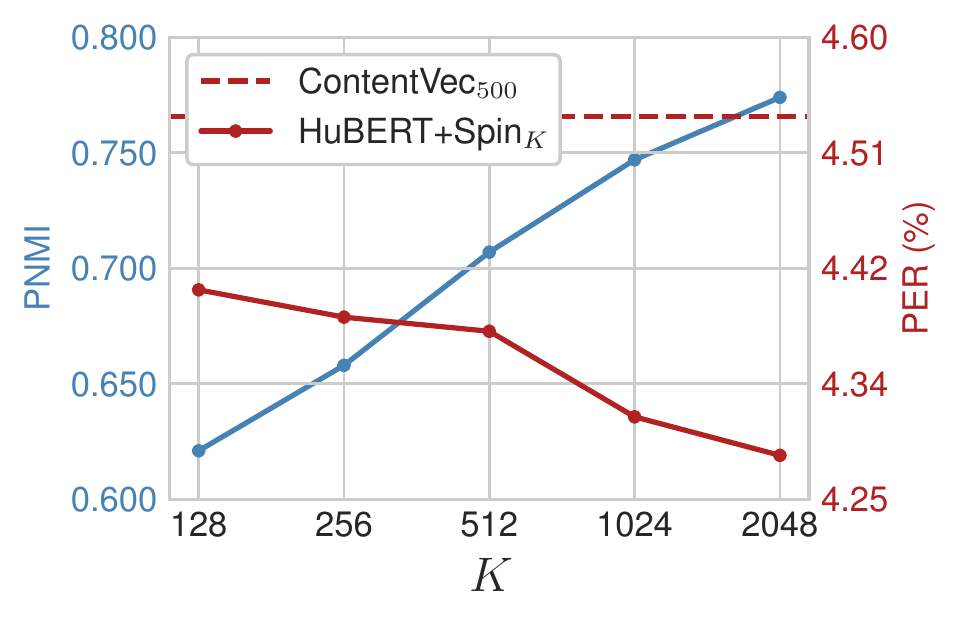}
         \vspace{-8pt}
         \caption{}
         \label{fig:pnmi-k}
     \end{subfigure}
     \begin{subfigure}[b]{0.33\linewidth}
         \centering
         \includegraphics[height=100pt]{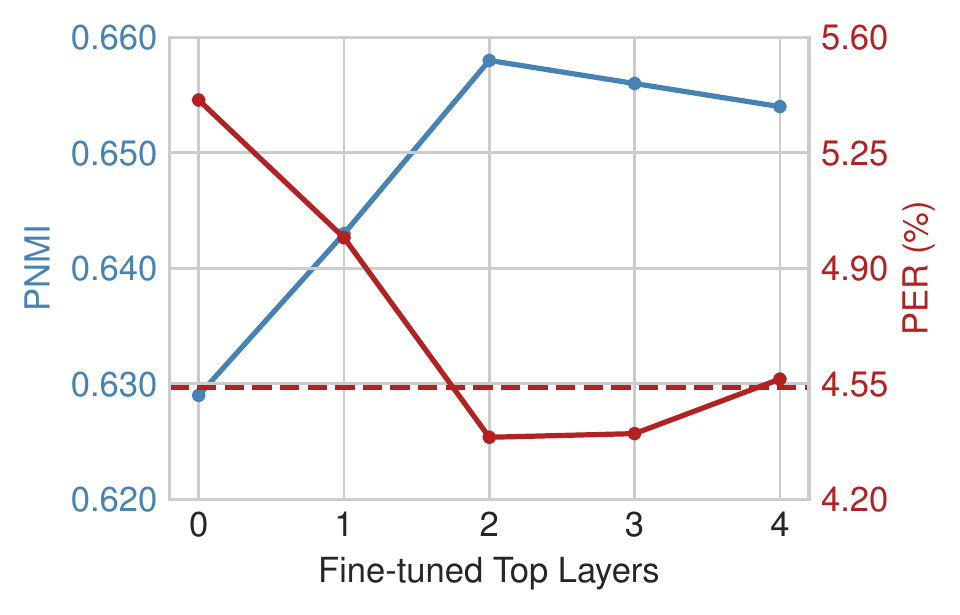}
         \vspace{-8pt}
         \caption{}
         \label{fig:layer-ft}
     \end{subfigure}
     \begin{subfigure}[b]{0.3\linewidth}
         \centering
         \includegraphics[height=100pt]{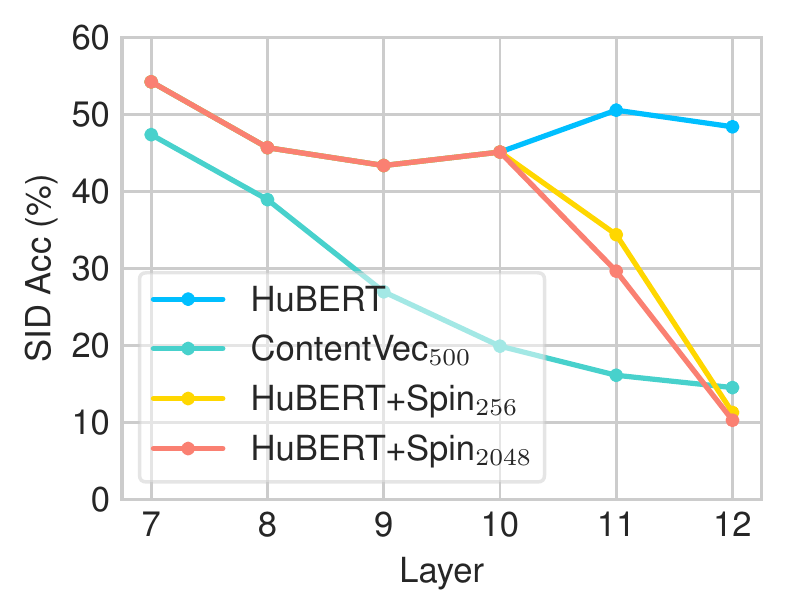}
         \vspace{-8pt}
         \caption{}
         \label{fig:layer-sid}
     \end{subfigure}
     \vspace{-8pt}
    \caption{
        PNMI and PER of HuBERT + \proposed~with different (a) codebook sizes and (b) fine-tuning layers.
        Fine-tuning zero layers in (b) denotes the HuBERT baseline.
        Results in (b) use $K=$ 256.
        (c) shows layer-wise speaker identification accuracy.
    }
    \vspace{-12pt}
    \label{fig:small-figs}
\end{figure*}
\begin{table}[t]
    \centering
    \caption{
        ABX error rates (\%) on the ZeroSpeech 2021 phonetic dev set~\cite{nguyen2020zero}.
        W and C before ``--'' denote within and across speakers.
        C and O after ``--'' denote clean and other corpus partitions.
        Only the layer with the lowest average score is reported for each model and is specified in column L.
    }
    \label{tab:abx}
    \vspace{-7pt}
    \begin{tabular}{@{~~}l@{~~}c@{~~~}c@{~~~}c@{~~~}c@{~~~}c@{~~~}c@{~~~}c@{~~}}
        \toprule
        Method & L & W-C & W-O & C-C & C-O & Avg \\
        \midrule
        Nguyen et al.~\cite{nguyen2022discrete} & -- & 3.26 & 3.81 & 4.00 & 5.91 & 4.25 \\
        Chorowski et al.~\cite{chorowski2021information} & -- & 2.95 & 3.54 & 4.50 & 7.05 & 4.51 \\
        \midrule
        HuBERT & 11 & 3.07 & 3.90 & 3.71 & 6.19 & 4.22 \\
        WavLM & 11 & 2.73 & 3.41 & 3.21 & 4.95 & 3.58 \\
        data2vec & 4 & 4.03 & 5.09 & 4.72 & 6.97 & 5.20 \\
        \midrule
        ContentVec\textsubscript{100} & 12 & 2.98 & 3.70 & 3.44 & 5.17 & 3.82 \\
        ContentVec\textsubscript{500} & 12 & 3.91 & 4.37 & 4.46 & 5.80 & 4.64 \\
        HuBERT + \proposed\textsubscript{2048} & 12 & \textbf{2.44} & \textbf{3.00} & \textbf{2.81} & \textbf{3.76} & \textbf{3.00} \\
        WavLM + \proposed\textsubscript{2048} & 12 & 2.75 & 3.33 & 3.24 & 4.17 & 3.37 \\
        \bottomrule
    \end{tabular}
    \vspace{-11pt}
\end{table}

\subsection{Acoustic Unit Discovery}
\label{subsec:zs}

This section inspects linguistic units captured in representations with Zero Resource Speech Benchmark~(ZeroSpeech) 2021~\cite{nguyen2020zero}.
The phonetic task measures how well speech representations distinguish between different phonemes via the ABX discrimination test~\cite{schatz2016abx}.
We report $K=$ 2048 since it performs the best in this task.
As shown in Table~\ref{tab:abx}, \proposed~boosts both models and surpasses the baselines, especially for HuBERT, surpassing prior art and reducing the average ABX error rate by a relative 29\%.
Although the performance gain for WavLM is minor, error rates of other corpus partitions are reduced, indicating that \proposed~helps WavLM in a noisier scenario.
The results directly demonstrate that \proposed~improves extracting phonemes.

\begin{table}[t]
    \centering
    \caption{
        Discrete unit quality.
        Cls Pur, Phn Pur, and PNMI denote cluster purity, phone purity, and phone-normalized mutual information~\cite{hsu2021hubert}.
        Only the layer with the highest PNMI is reported for each model and is specified in column L.
    }
    \label{tab:pnmi}
    \vspace{-7pt}
    \begin{threeparttable}
    \begin{tabular}{@{~~~}l@{~~~}c@{~~~~}c@{~~~}c@{~~~}c@{~~~}}
        \toprule
        Method & L & Cls Pur & Phn Pur & PNMI \\
        \midrule
        \multicolumn{5}{@{~~~}l}{\textbf{K-means Clustering} ($K=$ 256)} \\
        ~~~HuBERT & 7 & 0.154 & 0.639 & 0.630 \\
        ~~~WavLM & 11 & \textbf{0.178} & 0.624 & 0.640 \\
        ~~~data2vec & 4 & 0.173 & 0.652 & 0.630 \\
        ~~~ContentVec\textsubscript{100} & 12 & 0.169 & 0.650 & 0.643 \\
        ~~~ContentVec\textsubscript{500} & 8 & 0.154 & 0.639 & 0.629 \\
        ~~~HuBERT + \proposed\textsubscript{256} & 12 & 0.150 & 0.641 & 0.655 \\
        ~~~HuBERT + \proposed\textsubscript{2048} & 12 & 0.151 & \textbf{0.654} & \textbf{0.666} \\
        ~~~WavLM + \proposed\textsubscript{256} & 12 & 0.137 & 0.644 & 0.658 \\
        ~~~WavLM + \proposed\textsubscript{2048} & 12 & 0.153 & 0.650 & \textbf{0.666} \\
        \midrule
        \multicolumn{5}{@{~~~}l}{\textbf{Online Clustering} (\textbf{Codebook})} \\
        ~~~VQ-APC~\cite{chung2020vq-apc}$^{\spadesuit}$ & -- & 0.078 & 0.240 & 0.189 \\
        ~~~Co-training APC~\cite{yeh2022autoregressive}$^{\clubsuit}$ & -- & 0.089 & 0.308 & 0.294 \\
        ~~~HuBERT + \proposed\textsubscript{256}$^{\diamondsuit}$ & -- & \textbf{0.138} & 0.642 & 0.658 \\
        ~~~WavLM + \proposed\textsubscript{256}$^{\diamondsuit}$ & -- & 0.133 & \textbf{0.646} & \textbf{0.659} \\
        \bottomrule
    \end{tabular}
    \begin{tablenotes}[flushleft]
        \item{$^{\spadesuit}$}
            {\footnotesize 98 out of 512 codewords are utilized.}
            \vspace{-2pt}
        \item{$^{\clubsuit}$}
            {\footnotesize 164 out of 256 codewords are utilized.}
            \vspace{-2pt}
        \item{$^{\diamondsuit}$}
            {\footnotesize All 256 codewords are utilized.}
    \end{tablenotes}
    \end{threeparttable}
    \vspace{-12pt}
\end{table}

\begin{figure}
    \centering
    \begin{subfigure}[b]{0.495\linewidth}
        \centering
        \includegraphics[trim=7 0 3 0,clip,width=\linewidth]{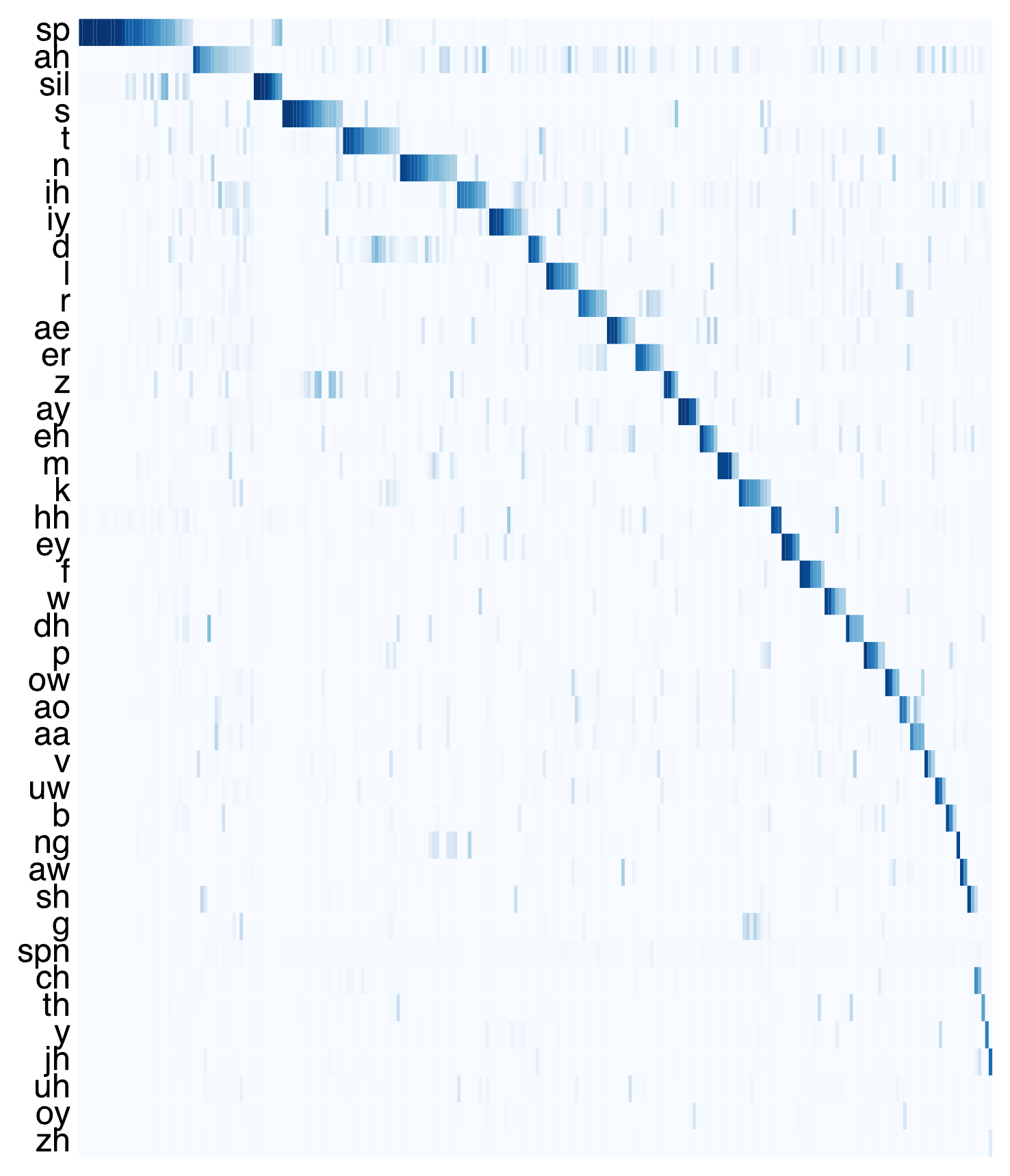}
        \vspace{-15pt}
        \caption{$K=$ 256}
        \label{fig:cp_256}
    \end{subfigure}
    \hfill
    \begin{subfigure}[b]{0.495\linewidth}
        \centering
        \includegraphics[trim=7 0 3 0,clip,width=\linewidth]{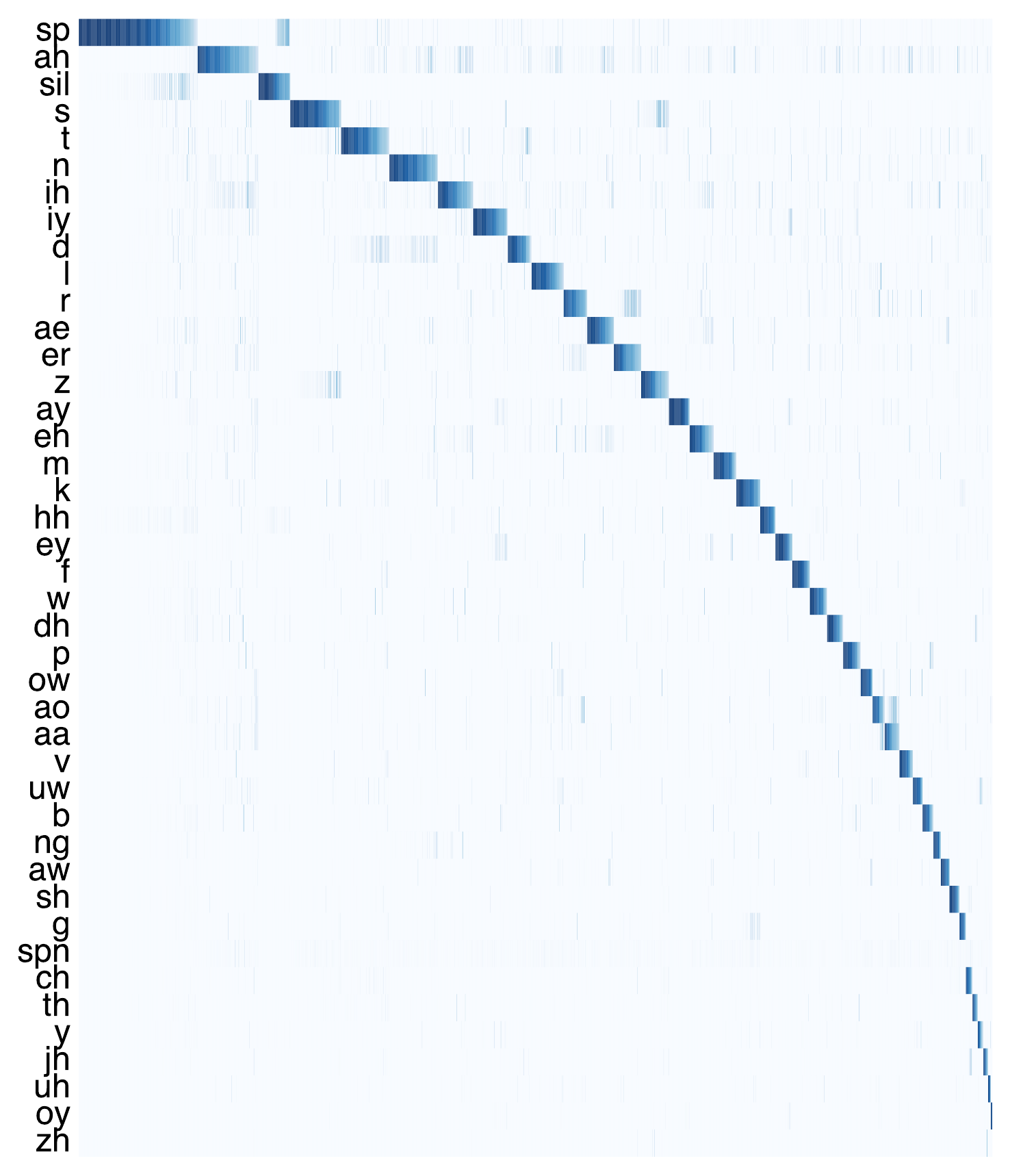}
        \vspace{-15pt}
        \caption{$K=$ 2048}
        \label{fig:cp_2048}
    \end{subfigure}
    \vspace{-18pt}
    \caption{
        $P(\text{\normalfont phone}|\text{\normalfont code})$ for HuBERT + \proposed.
        The vertical axes represent the phones sorted from high to low frequencies.
    }
    \vspace{-12pt}
    \label{fig:code-phone}
\end{figure}

\subsection{Discrete Unit Quality}
\label{subsec:cls-qual}

This section analyzes discrete acoustic unit quality to reveal the relationship between speech representations and phonemes.
We adopt three metrics proposed in HuBERT~\cite{hsu2021hubert}:
1) cluster purity measures the purity of each phoneme's associated discrete units; 2) phone purity measures the average phoneme purity within one class of discrete units; 3) phone-normalized mutual information (PNMI) measures the uncertainty reduction for the underlying phone when observing the codeword of a frame.
Higher values imply better performance.
K-means clustering is performed on a random 10 hours subset of LibriSpeech train-clean-100 split.
The discrete units are evaluated on the combination of LibriSpeech dev-clean and dev-other splits.
The offline clustering scores are averaged over three runs.

First, we cluster continuous representations into 256 clusters and report the layer with the highest PNMI, as shown in the upper part of Table~\ref{tab:pnmi}.
Independent of codebook sizes and pre-trained models, \proposed~outperforms all baselines in PNMI.
Increasing the codebook size in \proposed~improves all three metrics (\proposed\textsubscript{256} vs. \proposed\textsubscript{2048}), indicating that a larger codebook learns more fine-grained phoneme representations.

For online clustering (codebook learning), we compare the codebook in \proposed\textsubscript{256} with VQ-APC~\cite{chung2020vq-apc} and Co-training APC~\cite{yeh2022autoregressive}, where the latter two methods leverage codebook learning to improve content modeling.
We produce discrete units for \proposed~by taking $\arg\max$ over $p$ per frame.
In the lower part of Table~\ref{tab:pnmi}, codebooks in \proposed~achieve high PNMI compared with prior works.
Unlike prior methods, because of the constraint in Eq.~\ref{eq:q_star}, all learned codewords are utilized in \proposed.

Next, we visualize $P($phone$|$code$)$ in Fig.~\ref{fig:code-phone} to demonstrate the relation between learned codewords and phonemes.
Since the vertical axes are sorted by phoneme occurrence frequency in human speech, the figures show that \proposed~assigns more codewords to represent high-frequency phonemes.
Furthermore, because off-diagonal values of $K=$ 2048 are lower than those of $K=$ 256 (Fig.~\ref{fig:cp_2048} vs. \ref{fig:cp_256}), a larger codebook helps each code to focus on encoding one phoneme, consistent with phone purity in Table~\ref{tab:pnmi}.
Overall, \proposed~learns good discrete acoustic units and improves continuous representations in SSL models.

\subsection{Analysis}
\label{subsec:analysis}

\noindent\textbf{Codebook Size.}
In Fig.~\ref{fig:pnmi-k}, a larger codebook size improves discrete unit quality and PER, consistent with Sec.~\ref{subsec:zs} and \ref{subsec:cls-qual}.
Even when $K$ is only 128, \proposed~outperforms ContentVec.

\noindent\textbf{Fine-tuning Strategy.}
In Fig.~\ref{fig:layer-ft}, \proposed~surpasses HuBERT and ContentVec when fine-tuning two or three layers.
Moreover, fine-tuning two or three layers perform similarly, indicating that the choice of fine-tuning layers is flexible and robust.

\noindent\textbf{Speaker Invariance.}
We show each layer's performance in SUPERB speaker identification, where prediction heads are trained only with 50k updates.
In Fig.~\ref{fig:layer-sid}, independent of codebook sizes, \proposed~reduces identification accuracy to 10\% in the last layer, slightly lower than ContentVec, successfully removing speaker identity.

\section{Conclusion}
\label{sec:conclusion}

This paper proposes \proposed, a self-supervised fine-tuning method that improves content representations motivated by speaker disentanglement and the strong relationship between discrete unit quality and downstream performance.
We offer empirical evidence that the proposed method benefits various content-related tasks.
Although only applying to HuBERT and WavLM, \proposed~paves a new way to enhance speech representation models after pre-training at a very low cost.
Future works include applying \proposed~to other speech SSL models, introducing more complex data augmentation to improve robustness, and extending to pre-train networks from scratch.

\bibliographystyle{IEEEtran}
\bibliography{mybib}

\end{document}